# Visual Indeterminacy in GAN Art[1]

Aaron Hertzmann

Adobe Research

**Abstract**

This paper explores *visual indeterminacy* as a description for artwork created with Generative Adversarial Networks (GANs). Visual indeterminacy describes images which appear to depict real scenes, but, on closer examination, defy coherent spatial interpretation. GAN models seem to be predisposed to producing indeterminate images, and indeterminacy is a key feature of much modern representational art, as well as most GAN art. It is hypothesized that indeterminacy is a consequence of a powerful-but-imperfect image synthesis model that must combine general classes of objects, scenes, and textures.

## 1. Introduction

The first widespread trend in machine learning-based artwork is the use of Generative Adversarial Networks (GANs) [1] [2]. Artists using them include Refik Anadol, Robbie Barrat, Sofia Crespo, Mario Klingemann, Trevor Paglen, Jason Salavon, Helena Sarin, and Mike Tyka; GAN-based work has been auctioned at both Christie's and Sotheby's. These artists work in very different ways and in different art contexts, but each has produced some work that shares a common GAN aesthetic: images that seem realistic but yet somehow unrecognizable.

This paper explores how the concept of *visual indeterminacy* provides a useful description for typical GAN artwork. Visual indeterminacy is a major theme in modern and contemporary art, and GANs seem to naturally produce indeterminate images. This paper hypothesizes that this is a consequence of the development of realistic image-based generative models, and suggests future directions for the development of models with explicit representation of visual uncertainty.

Visual indeterminacy, a term coined by Pepperell [3], describes imagery that appears at first to be coherent and realistic, but that defies consistent spatial interpretation. Often, the initial appearance of an image invites the viewer to investigate further, but the image confounds explanation. For some images, this investigation leads to an "Aha!" moment, where the viewer understands the structure of an image [4], e.g., they see a vivid 3D object where there had been abstract 2D shapes. This moment is pleasurable because some understanding has been gained, but the image may also become less interesting as result. "Visual indeterminacy" describes images where the "Aha!" moment never happens, and the image continues to invite investigation (Figure 1). In short, visual indeterminacy occurs in a "seemingly meaningful visual stimulus that denies easy or immediate identification" [5]; it is the "lack but promise of semantic stability" [6].

---

[1] Preprint version; final version to appear in *Leonardo* / SIGGRAPH Art Papers. A previous version of this paper was presented at the NeurIPS 2019 creativity workshop.

Note that a blurry image is not technically indeterminate, just ambiguous. Likewise, an image that admits two contradictory but self-consistent interpretations (such as Rubin's vase illusion or a Giuseppe Arcimboldo painting) is bistable, not indeterminate.

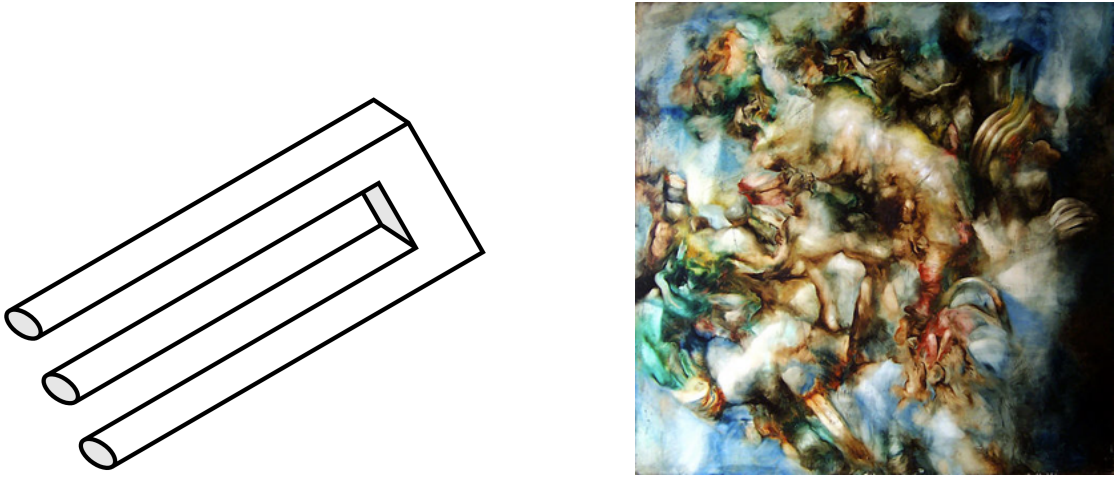

**Figure 1.** Indeterminate images. **Left**: the "impossible trident" appears to be a simple geometric object, but no matter where the eye fixates, no coherent 3D geometric interpretation is possible. **Right**: *Succulus*, Robert Pepperell, oil on canvas, 2005, painted specifically to be indeterminate (© Robert Pepperell.)

As surveyed by Gamboni [7], visual indeterminacy has been present in art since cave painting, but became particularly valued during the Modern art era. Cubism is an early modern example of indeterminate pictorial arrangements, about which Gombrich wrote "each hypothesis we assume will be knocked out by a contradiction elsewhere" [8]. As a current example, Gerhard Richter describes his aims: "We only find paintings interesting because we always search for something that looks familiar to us. … And usually we do find those similarities and name them: table, blanket, and so on. When we do not find anything, we are frustrated and that keeps us excited and interested …" [5, 9]

## 2. Visual Indeterminacy of GAN Art

Although GAN models can create completely realistic images, these images are not typically shown as artwork. Most GAN art appears ambiguous or visually indeterminate in some way; Mario Klingemann coined the term "Francis Bacon effect" to describe the surreal appearance of early GAN art [2].

This could describe his prize-winning "The Butcher's Son," which appears at first to be a grotesque nude portrait, but precludes clear identification of facial details or where exactly the chest and arms bleed into the background. Trevor Paglen's Adversarial Evolved Hallucinations, evoke objects in different categories, like teeth and gums in "False Teeth," or a roadside in "Highway of Death," without resolving into individual details. Each image in Jason Salavon's "Narrative Walk" and "Narrative Frame" works comes from a GAN trained on either illuminated manuscripts, newspapers, or websites, and each appears to include printed text from one of these classes, but the individual text and design elements cannot be read. Sofia Crespo's "Neural Zoo" appears at first to be biological imaging and photography, but shapes often blend and merge in uninterpretable ways. Refik Anadol's "Machine Hallucinations" and Mike Tyka's "EONS" videos show sequences of synthesized architecture and landscapes. These videos include imagery that is usually recognizable as buildings or landscapes, but specific details are ambiguous and hard to interpret.

Some GAN outputs appear as simply realistic imagery with simple visual artifacts, rather than being indeterminate. "Edmond de Belamy" just seems blurry. Many GAN images of people and animals appear grotesque, whether indeterminate or not.

GANs seem predisposed to indeterminate, intriguing imagery. This can be seen by experimenting with the General model in Artbreeder [10], formerly Ganbreeder (Figure 2). The General model allows users to generate new images from BigGAN [11]; images created this way have themselves been exhibited as art [12]. Just a few minutes exploring these models quickly produces uncanny and indeterminate imagery. That GANs seem predisposed to this sort of indeterminate art is also evidenced by the proliferation of early art often derided as "push-button GAN images" [1] [2]. Perhaps it seems easy to superficially simulate contemporary representational art with GANs because they produce so many different indeterminate images.

### 3. Why do GANs create indeterminate images?

This section hypothesizes a way to understand why GANs produce indeterminate imagery. At a high level, GANs are designed to generate novel realistic images that look as though they could have come from the real world. In order to this, they must somehow represent the different types of scenes, objects, and textures in the real world, and be able to compose and rearrange them in realistic ways. Algorithms that merely "cut-and-paste" elements from training images [13] [14] cannot model the continuous range of variations in the world, whereas composing 3D models of the real world would be too difficult to learn at present.

Recent studies [15, 16] suggest that GANs solve this problem by arranging objects into scenes, creating the objects, and then adding color, texture, and lighting to illustrate these objects, all in pictorial space. These arrangements and textures are not discrete but continuous—objects need not have distinct boundaries in the image, and need not be complete. This can yield impossible combinations of object identities, locations, and textures that all bleed into one another. That is, the object creation and texturing steps do not operate on separate, distinct objects, and object parts and textures can merge and blend across objects, like filling in a coloring book where none of the outlines are closed and none of the shapes are quite right, or putting together puzzles pieces from different puzzles. This produces visual indeterminacy.

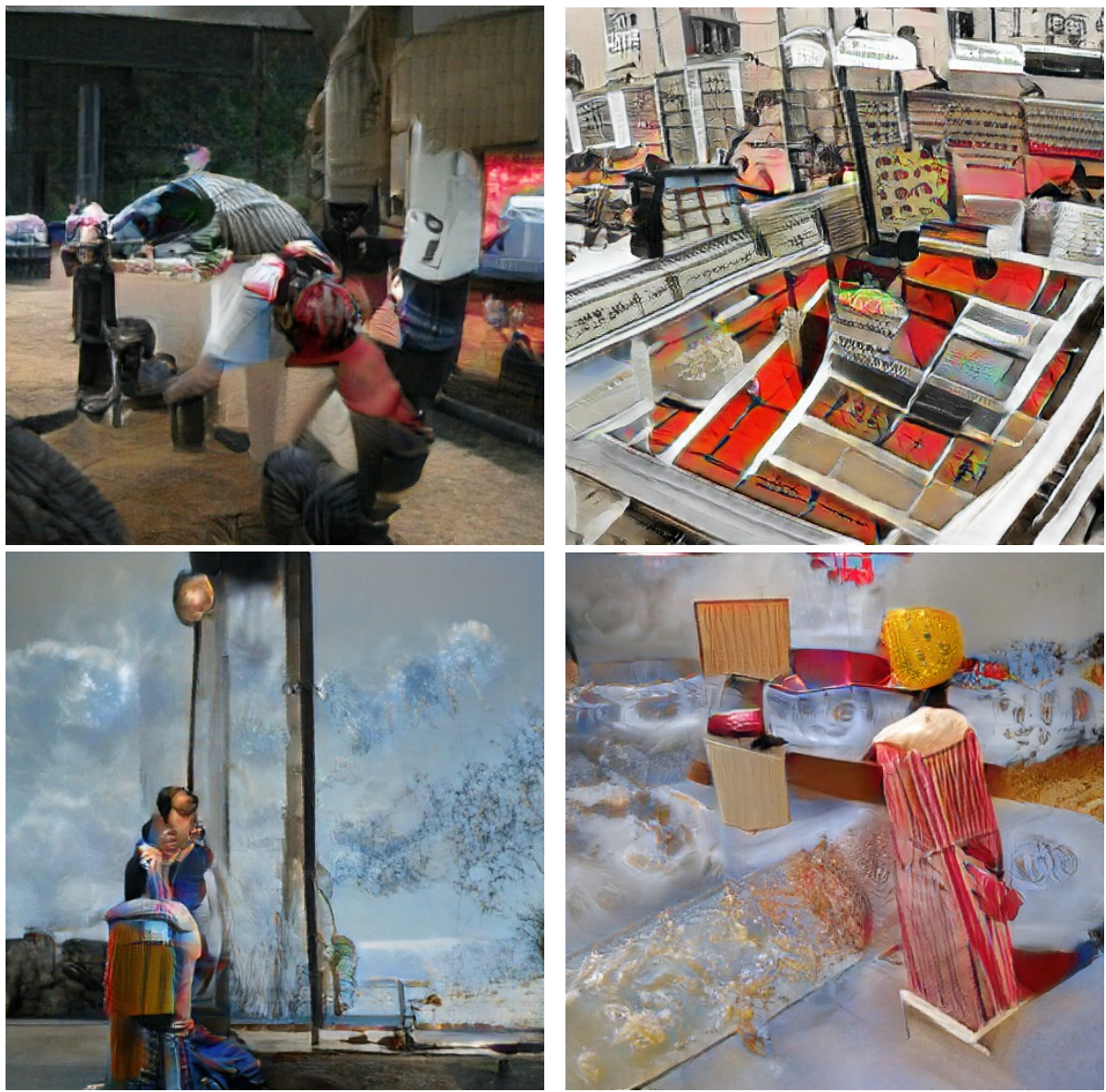

**Figure 2.** Indeterminate images from BigGAN, created with Artbreeder.com (Public domain imagery by Artbreeder users caincaser; guidoheinze; jeffgiddens; suddenwoven/jasongraf)

It may be that the ability to create indeterminacy is not just an accident specific to GANs, but a consequence of any powerful-but-imperfect generative model. A generative image model needs to be able to combine objects and textures in ways not present in the training images, and any realistic image model that has similar compositional properties will misfire, producing *nearly* realistic combinations. Realistic image modeling is such a difficult problem that indeterminate imagery is to be expected from many future general-purpose neural generative models.

Eventually, generative networks may get so good that they rarely, if ever, produce unrealistic images. Perhaps there is an "Uncanny Ridge," along which generators are only just good enough to produce a diverse set of intriguingly indeterminate images, and past which the outputs are less and less surreal (Figure 3). Once generators surpass this Uncanny Ridge, artists will need to "break" the models to produce imagery that does not just appear to be an ordinary photograph. This will also be necessary, regardless, as GAN fatigue sets in [2]. Already, the most visually innovative GAN artists actively manipulate and "break" their GAN models, such as in Helena Sarin's amplification of network artifacts [17] and Klingemann's "Neural Glitch" work [18]. Further exploration of these sorts of generative manipulations will be important for artists to develop GAN art beyond mere novelty-exploration [1].

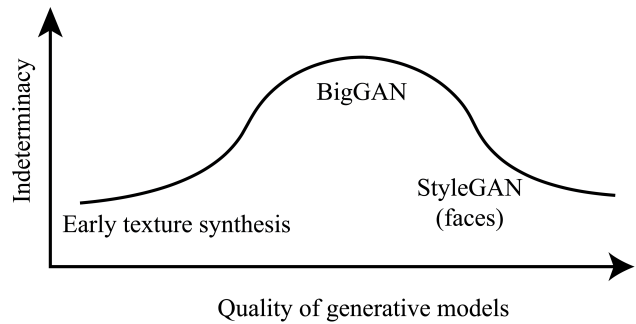

**Figure 3.** The Uncanny Ridge. To create indeterminate imagery, models must be good enough to produce realistic novel imagery, but not so good that all their outputs look like ordinary photographs.

**4. Towards the Convergence of Perception and Aesthetics**

GANs arose indirectly from past research in neuroscience and computer vision. Will we one day be able to close the loop between image synthesis and perception?

Several recent theories connect indeterminacy to perceptual neuroscience; see [6] [19] for reviews, and [20] for a discussion of visual ambiguity in terms of perceptual posterior probability. Specifically, Muth and Carbon [6] explain visual indeterminacy in terms of predictive coding. Predictive coding is a prominent neuroscience theory for implementing perceptual uncertainty while explaining various observations from neurological studies. In predictive coding, neural representations for different aspects of image interpretation make predictions for other aspects. For example, if the scene recognition module interprets a scene as containing a face, then it sends eyes and nose predictions to modules for specific locations. Conversely, the presence of eyes will send prediction of a face to an overall scene recognition module. If all predictions are met, then no more predictions are sent and the process stabilizes. Prediction error signals are generated when predictions are not met, leading to updated predictions. Indeterminacy occurs when there is no stable resolution to this process.

Natural image models, vision neuroscience, and image synthesis have long fed into each other. Discoveries about the visual cortex [21] led to natural image statistics analysis [22] [23], which led to texture synthesis algorithms [13] [24] which led to style transfer algorithms [25] [14] [26]. At the same time, cortical modeling also led to deep convolution networks [27], which led to GANs and trained discriminative networks, which, in turn, have led to improved neuroscience models [28]. Can aspects of aesthetic experience be understood with the same models, and could this lead to more powerful artistic tools?

Ideally, a recognition model would approximate human perceptual uncertainty by providing a posterior probability distribution over interpretations. Optimization against this model would give artists more precise control over perceptual uncertainty in images, for example, to produce images with specific types of ambiguity and visual indeterminacy.

This would provide artists with higher-level controls to explore artistic creation. It could also provide a richer testbed to develop perceptual theories of aesthetic experience, rather than using hand-crafted artworks as in [3] [4].

Visual Indeterminacy in Generative Neural Art8. E. Gombrich, *Art and Illusion: A Study in the Psychology of Pictorial Representation*, (Princeton: Princeton University Press, 1960).

9. R. Storr, *Gerhard Richter: Doubt and Belief in Painting*, (New York: Museum of Modern Art, 2003).

10. J. Simon, "Artbreeder," https://artbreeder.com.

11. A. Brock, J. Donahue and K. Simonyan, "Large Scale GAN Training for High Fidelity Natural Image Synthesis," in *Proc. ICLR*, 2019.

12. J. Bailey, "Why is AI art so complicated?," 27 March 2019. https://www.artnome.com/news/2019/3/27/why-is-ai-art-copyright-so-complicated.

13. A. Efros and T. Leung, "Texture Synthesis by Non-Parametric Sampling," in *ICCV*, 1999.

14. A. Hertzmann, C. E. Jacobs, N. Oliver, B. Curless and D. H. Salesin, "Image Analogies," in *Proc. SIGGRAPH*, 2001.

15. D. Bau et al., "GAN dissection: Visualizing and Understanding Generative Adversarial Networks," in *Proc. ICLR*, 2019.

16. C. Yang, Y. Shen and B. Zhou, "Semantic Hierarchy Emerges in Deep Generative Representations for Scene Synthesis," arxiv.org:1911.09267, 2019.

17. H. Sarin, "Checkerboard artifacts amplification series," 19 August, 2019. https://twitter.com/glagolista/status/1121957807606194183.

18. M. Klingemann, "Neural Glitch," *Issues in Science and Technology,* vol. 36, no. 2, 28 October 2020.

19. S. Van de Cruys and J. Wagemans, "Putting reward in art: A tentative prediction error account of visual art," *i-Perception,* vol. 2, no. 9, 2011.

20. A. Hertzmann, "Non-Photorealistic Rendering and the Science of Art," in *Proc. NPAR*, 2010.

21. D. H. Hubel and T. N. Wiesel, "Receptive fields and functional architecture of monkey striate cortex," *J. Physiology,* vol. 195, no. 1, 1968.

22. D. J. Field, "Relations between the statistics of natural images and the response properties of cortical cells," *J. Opt. Soc. Am. A,* vol. 4, no. 12, pp. 2379-2394, 1987.
Page 7 of 8